\documentclass[runningheads]{llncs}

\usepackage{xcolor}
\usepackage[T1]{fontenc}

\usepackage{algorithm}
\usepackage{algorithmic}
\usepackage{graphicx,verbatim}
\usepackage{amsmath}
\usepackage{booktabs}

\usepackage{amssymb}
\usepackage{pifont} 
\usepackage{svg}
\usepackage{multirow}  
\usepackage{booktabs}  
\usepackage{graphicx}  
\usepackage{colortbl}
\usepackage{xcolor}

\newcommand{\redcirc}{\textcolor{red}{\bullet}}
\newcommand{\greencirc}{\textcolor{green!60!black}{\bullet}}
\usepackage{soul}
\newcommand{\se}[1]{\ensuremath{_{\scriptsize\pm#1}}}

\begin{document}
\title{Worst-Group Equalized Odds Regularization for Multi-Attribute Fair Medical Image Classification}
\titlerunning{Worst-Group EO for Multi-Attribute Fairness}
%

\author{
Nikhil Cherian Kurian\inst{1} \and
Victor Caquilpan Parra\inst{1} \and
Abin Shoby\inst{1} \and
Luke Whitbread\inst{1} \and
Lauren Oakden-Rayner\inst{1} \and
Robert Vandersluis\inst{2} \and
Jessica Schrouff\inst{2} \and
Lyle J. Palmer\inst{1} \and
Mark Jenkinson\inst{1}
}

\authorrunning{Kurian et al.}

\institute{
Australian Institute for Machine Learning, Adelaide University, Adelaide, Australia\\
\email{\{nikhil.kurian, abin.shoby, luke.whitbread, lyle.palmer, mark.jenkinson\}@adelaide.edu.au}\\
\email{victor.caquilpan@gmail.com, laurenoakdenrayner@gmail.com}
\and
GlaxoSmithKline (GSK)\\
\email{\{robert.x.vandersluis, jessica.v.schrouff\}@gsk.com}
}
\maketitle              
%
\begin{abstract}
\text Diagnostic performance in medical AI varies systematically across demographic groups, yet subgroup AUC can mask clinically important disparities. At a fixed inference-time operating point, some groups may exhibit over-diagnostic behaviour, characterized by elevated true and false positive rates, while others show under-diagnostic patterns with reduced true and false positive rates. These opposing tendencies can cancel in aggregate AUCs while producing meaningful inequities in clinical decision-making. Motivated by the need to assess and mitigate such disparities at the operating point and across multiple demographic attributes simultaneously, we propose a worst-group equalized-odds margin regularizer. The proposed regularizer explicitly targets subgroup-level deviations on both the true positive and false positive sides at inference. At each update, the method identifies subgroups defined by explicit demographic attributes (e.g., age, sex, and race) that exhibit the most extreme margin deviations and applies a unified penalty, enabling fairness optimization across multiple demographic axes without requiring explicit intersectional constraints. Across two medical imaging datasets in realistic multi-label settings, our method consistently reduces disparities in Equalized Odds and Equalized Opportunity with minimal impact on AUC, preserving diagnostic performance while improving fairness.


\end{abstract}
\section{Motivation and Background}
Fairness is a critical concern in medical AI, as diagnostic performance often varies systematically across demographic groups~\cite{xu2024addressing,mehrabi2021survey}. At inference time, subgroups defined by age, sex, or race/ethnicity can exhibit markedly different error profiles—for example, younger females may have lower true positive rates (TPR) than older males for the same condition~\cite{seyyed2020chexclusion}. Such disparities can be obscured when subgroup AUCs appear similar: at a fixed global threshold, some groups are effectively over-diagnosed (high TPR and false-positive rate, FPR), while others are under-diagnosed (low TPR and FPR), leading opposing behaviours to cancel out in aggregate~\cite{glocker2023algorithmic}. In radiology benchmark studies, race-stratified AUCs for White and Black patients are often comparable, yet at a global operating point Black patients show reduced sensitivity while White patients experience higher false-positive rates \cite{seyyed2020chexclusion,glocker2023algorithmic}. Consequently, fairness is better assessed using Equalized Odds (EO) and Equalized Opportunity (EOpp), which explicitly capture TPR/FPR disparities that AUC-based summaries can miss~\cite{hardt2016equality}.

Despite this, many existing mitigation methods—such as GDRO \cite{DRO2020}, JTT ~\cite{JTT2021}, or FairBatch~\cite{fairbatch2021}—reweight, resample, or modify log-loss, an objective primarily intended to reduce average predictive risk which may fail to capture localized subgroup disparities~\cite{IncorporatingRatherEliminating2025}. Moreover, most prior work considers fairness relating to a single demographic attribute, leaving disparities along other axes unaddressed or even exacerbated~\cite{kearns2018preventing}. Approaches that address multiple attributes together (e.g., FairRead~\cite{gao2025fairread}, FCRO~\cite{deng2023fairness}) often have to adopt coarsened subgroup definitions across all attributes to preserve statistical power within intersectional groups. Representation-level debiasing methods that enforce demographic invariance or suppress mutual information~\cite{kearns2018preventing,oakden2020hidden} also struggle in multi-attribute settings: they become unstable for small subgroup sizes, and removing demographic signal independently for each attribute can overcompensate, eliminating task-relevant information and worsening the fairness–accuracy trade-off. Most such approaches are evaluated only in simplified binary classification settings, limiting their relevance to the multi-label scenarios common in clinical imaging~\cite{mehrabi2021survey}.

In this work, we study fairness across multiple demographic subgroups in a realistic multi-label diagnostic setting. We introduce a worst-group equalized-odds margin regularizer that targets operating-point level subgroup disparities using threshold-free logit margins. Beyond standard log-loss—which primarily improves AUC through re-ranking—our bi-directional regularizer penalizes both over-diagnosis and under-diagnosis. Crucially, our definition of the “worst” subgroup is attribute inclusive: at each update, we search across all demographic groups (e.g., age bins, race categories, sex groups) and identify the subgroup  exhibiting the largest potential deviation in TPR or FPR, without maintaining separate constraints per attribute. This provides a unified mechanism to address fairness across multiple demographic axes. Empirically, we observe consistent improvements in EO and EOpp with minimal changes in overall AUC, reducing subgroup disparities without incurring substantial accuracy losses or improving fairness for one subgroup at the expense of worsening it for others—a behaviour commonly observed in classical fairness–accuracy trade-offs~\cite{menon2018cost,mehrabi2021survey,gao2025fairread}.

\section{Method}
EO defines fairness via parity of true and false positive rates across demographic groups at a fixed inference-time operating point~\cite{xu2024addressing}. Cross-entropy training improves overall score separation and often benefits groups with already favorable TPR and FPR at test time, but does not explicitly penalize worst-case subgroup deviations, allowing large EO violations to persist despite strong average performance. We therefore introduce a Worst-Group Equalized Odds regularizer that targets TPR and FPR deviations by improving the separation of worst-group positives from negatives, and vice versa, using a threshold-free differentiable logit margin-based objective.

\subsection{Problem Formulation}

Consider a binary classification dataset $\mathcal{D}=\{(\mathbf{x}_i, y_i, \mathbf{a}_i)\}_{i=1}^N$, where $\mathbf{x}_i \in \mathcal{X}$ is the input (e.g., a medical image), $y_i \in \{0,1\}$ is the label, and $\mathbf{a}_i = (a_i^{\text{age}}, a_i^{\text{sex}}, a_i^{\text{race}})$ denotes the demographic attributes of sample $i$. 
For each attribute (e.g., age, sex, race), we define a set of demographic subgroups of interest, and collect all such subgroups into a unified set $\mathcal{G}$. Each subgroup $g \in \mathcal{G}$ corresponds to a single attribute--value pair (e.g., \emph{female} for sex, \emph{White} for race, or \emph{age} $\geq 60$). 
For notational convenience, we write $g_i = g$ to indicate that sample $i$ belongs to subgroup $g$ according to its attributes $\mathbf{a}_i$; a sample may satisfy this relation for multiple subgroups. We use a neural network $f_\theta:\mathcal{X} \rightarrow \mathbb{R}$ that outputs logits $\ell_i = f_\theta(\mathbf{x}_i)$, which are mapped to probabilities via a link function such as the sigmoid, $p_i = \sigma(\ell_i) = \frac{1}{1 + e^{-\ell_i}}$.

\subsection{Worst Sub-Group Identification}

To address fairness violations, we dynamically identify the worst-performing subgroups (agnostic to the attribute as mentioned before) within each mini-batch. For positives ($y_i = 1$), we seek the demographic subgroup $g_{\min}^{[+]}$, with the lowest mean predicted probability (as a surrogate of TPR - indicating under-diagnosis risk), while for negatives ($y_i = 0$), we identify the subgroup $g_{\max}^{[-]}$, with the highest mean probability (as a surrogate of FPR - indicating over-diagnosis risk). We compute per-group average probabilities by $\mu_g^{[+]} = mean\{p_i \mid y_i = 1, g_i = g\}$ and $\mu_g^{[-]} = mean\{p_i \mid y_i = 0, g_i = g\}$. The worst-performing subgroups are then identified as $g_{\min}^{[+]} = \arg\min_{g} \mu_g^{[+]}$ for positives and $g_{\max}^{[-]} = \arg\max_{g} \mu_g^{[-]}$ for negatives.

\subsection{Log-Sum-Exp Margin Regularization}
Our goal is to penalize cases in which the lowest-scoring positive sample from the worst subgroup $g_{\min}^{[+]}$ is scored below all negative samples and conversely where the highest scoring negative sample from the worst subgroup $g_{\max}^{[-]}$ is scored above all positive samples
This encourages separation between positive and negative samples from the worst-group, reducing under-diagnosis and over-diagnosis, respectively. Ideally we want

\begin{align}
\min_{\substack{m: y_m=1, \\ g_m=g_{\min}^{[+]}}} \ell_m > \max_{\substack{n: y_n=0}} \ell_n \qquad
\max_{\substack{m: y_m=0, \\ g_m=g_{\max}^{[-]}}} \ell_m < \min_{\substack{n: y_n=1}} \ell_n \label{eq:tpr_fpr_constraint}
\end{align}

\begin{figure}[t]
    \centering
    \includegraphics[width=\linewidth,height=5.2cm]{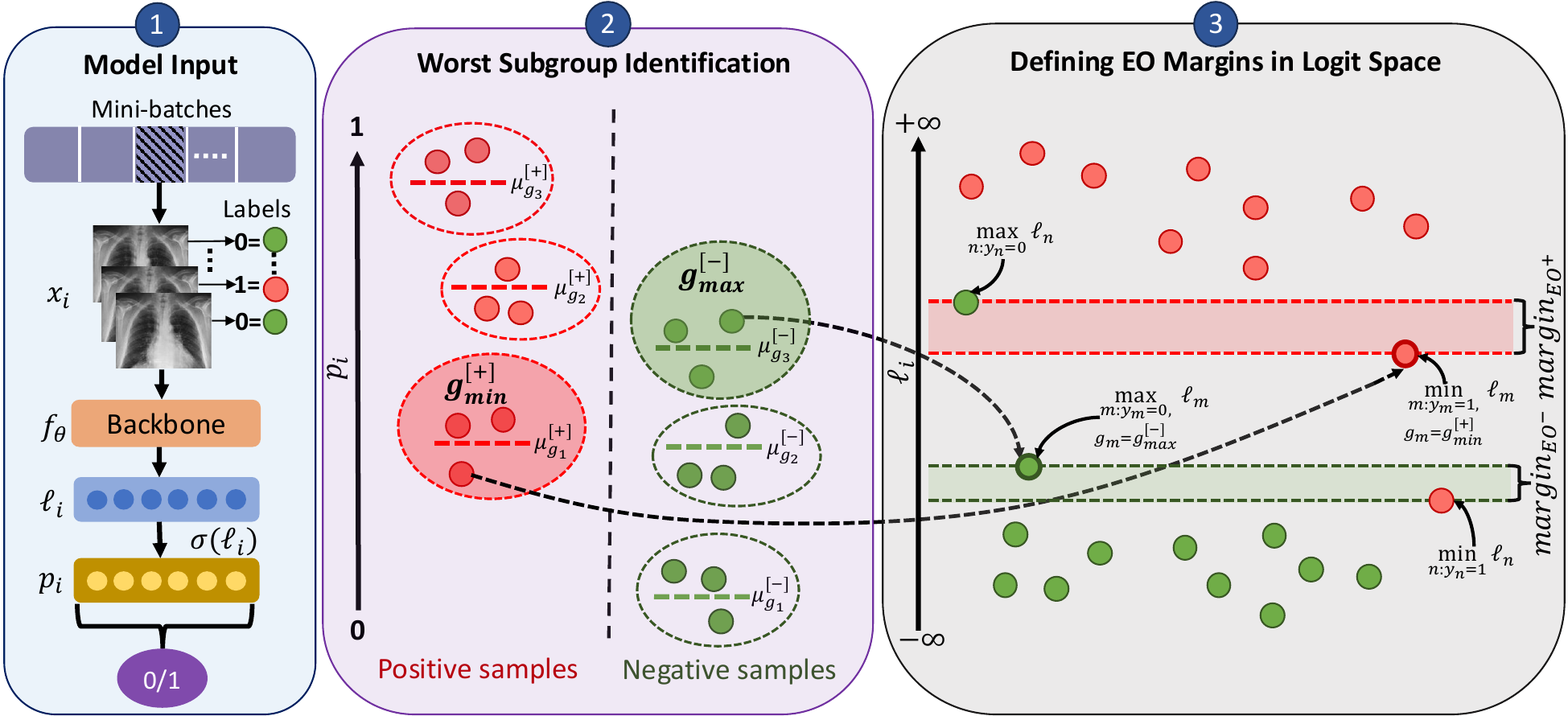}
\caption{
\textbf{Defining EO Margins:} (1) Aggregate samples by label:
$\redcirc$ (positive), $\greencirc$ (negative).
(2) Identify $g_{\min}^{[+]}$ as the positive subgroup with lowest $\mu^{[+]}_{g_i}$.
(3) Define $margin_{\mathrm{EO}^{+}}$ as the separation between the worst sample in
$g_{\min}^{[+]}$ and the worst negative sample;
$margin_{\mathrm{EO}^{-}}$ is defined analogously for the worst negative sample in $g_{\max}^{[-]}$.}

     \label{fig1:main_approach}
\end{figure}

In Eq.~(\ref{eq:tpr_fpr_constraint}), $m$ and $n$ index individual samples within the current mini-batch.
The index $m$ is restricted to samples from the identified worst subgroup (positives with $g_m=g_{\min}^{[+]}$ in the left constraint, negatives with $g_m=g_{\max}^{[-]}$ in the right), whereas $n$ ranges over all samples with the opposite label in the batch.
Thus, each inequality compares the worst-case logit in the worst subgroup (a $\min$ over positives or a $\max$ over negatives) against the most confusable extreme of the opposite class, enforcing score separation to reduce under-diagnosis (potential TPR disparity) and over-diagnosis (potential FPR disparity). Fig.~\ref{fig1:main_approach} summarizes this formulation setup.

However, these hard min/max operators depend on single extreme samples and are non-differentiable, which makes them unsuitable for gradient-based optimization. We therefore replace the hard extrema in Eq.~(\ref{eq:tpr_fpr_constraint}) with a smooth relaxation based on the log-sum-exp (LSE) operator, which preserves the focus on worst-case behavior while enabling stable gradient-based training:
\begin{align}
\operatorname{LSE}_{\max}(\{x_i\}_{i\in\mathcal{I}})
&= \log \sum_{i\in\mathcal{I}} e^{x_i}
\approx \max_{i\in\mathcal{I}} x_i \\
\operatorname{LSE}_{\min}(\{x_i\}_{i\in\mathcal{I}})
&= -\log \sum_{i\in\mathcal{I}} e^{-x_i}
\approx \min_{i\in\mathcal{I}} x_i
\label{eq:lse}
\end{align}
where $\mathcal{I}$ denotes the index set of samples over which the aggregation is performed (e.g., all positives in the identified worst subgroup or all negatives in the mini-batch), with cardinality $|\mathcal{I}|$. 

Beyond enabling differentiability, this relaxation allows samples close to the extremum to contribute to the penalty instead of relying solely on a single extreme value. Such poorly performing samples may belong to multiple demographic subgroups, and this behavior supports improved mitigation of intersectional fairness violations.

We define LSE-based margins to quantify worst-group separation:
$\text{margin}_{\text{EO}^+}
=\operatorname{LSE}_{\max}(\{\ell_i : y_i=0\})
-\operatorname{LSE}_{\min}(\{\ell_i : y_i=1,\, g_i=g_{\min}^{[+]}\})$
captures TPR-side violations by comparing worst-group positives to negatives, while
$\text{margin}_{\text{EO}^-}
=\operatorname{LSE}_{\max}(\{\ell_i : y_i=0,\, g_i=g_{\max}^{[-]}\})
-\operatorname{LSE}_{\min}(\{\ell_i : y_i=1\})$
captures FPR-side violations by comparing worst-group negatives to positives. Therefore,

\begin{align}
\text{margin}_{\text{EO}^+} &= \log \sum_{y_i=0} e^{\ell_i} + \log \sum_{y_i=1, g_i=g_{\min}^{[+]}} e^{-\ell_i} \label{eq:tpr_margin} \\
\text{margin}_{\text{EO}^-} &=  \log \sum_{y_i=0, g_i=g_{\max}^{[-]}} e^{\ell_i} + \log \sum_{y_i=1} e^{-\ell_i}  \label{eq:fpr_margin}
\end{align}

\subsection{Hinge Loss and Final Objective}

We apply a hinge mechanism to penalize only positive margins (violations) to create our training objectives: $\mathcal{L}_{\text{EO}^{+}} = \max(0, \text{margin}_{\text{EO}^+})$ and $\mathcal{L}_{\text{EO}^{-}} = \max(0, \text{margin}_{\text{EO}^-})$.

If any required sample pool is empty within a mini-batch (e.g., no negatives), the corresponding loss term is set to zero to avoid numerical instability. The complete training objective combines the base BCE loss with both fairness regularizers as $\mathcal{L} = \mathcal{L}_{\text{base}} + \lambda_{\text{EO}^{+}} \mathcal{L}_{\text{EO}^{+}} + \lambda_{\text{EO}^{-}} \mathcal{L}_{\text{EO}^{-}}$. Further, in case of multi-label settings, $\mathcal{L}_{\text{base}}$ and $\mathcal{L}_{\text{EO}}$ losses will be averaged across the classes. 

Here, ${L}_{\text{EO}^{+}}$ targets disparities in positive predictions, which correspond to TPR-side EO violations and ${L}_{\text{EO}^{-}}$ targets disparities in negative predictions, which correspond to FPR-side EO violations. The approach naturally induces a fairness-accuracy trade-off \cite{menon2018cost} controlled by $\lambda_{\text{EO}^{+}}$ and $\lambda_{\text{EO}^{-}}$, which should be tuned via fairness-aware validation metrics such as EO gap or worst-group performance~\cite{kearns2018preventing}. This formulation provides several advantages. By identifying worst subgroups per mini-batch and applying LSE-based margins, we achieve differentiable worst-group optimization while maintaining computational efficiency. The regularizer directly targets worst-case EO violations across demographic groups in a multi-attribute setting.
 Unlike hard min/max operations that concentrate gradients on outliers, LSE margins distribute gradient flow across samples, improving convergence stability. The regularization operates during training without requiring post-hoc threshold adjustment.

 Although AUC metric can mask clinically relevant disparities at the operating point, our regularizer can be interpreted as localized AUC optimization: $\mathcal{L}_{\text{EO}^{+}}$ strengthens the ranking of weak positives from the worst-performing subgroup against all negatives, while $\mathcal{L}_{\text{EO}^{-}}$ induces an analogous ranking objective under label inversion, concentrating ranking pressure on score regions responsible for TPR and FPR disparities.

\subsection{Datasets and Preprocessing}

We use two medical imaging datasets with comprehensive demographic annotations. \textbf{RNFL-OCT}~\cite{Glaucoma2024} is an ophthalmic dataset comprising 3,300 retinal nerve fiber layer thickness (RNFLT) 2D projection maps, with a single binary glaucoma classification target (52\% positive). The dataset poses challenges for multi-attribute fairness analysis due to its limited scale. While prior studies have examined fairness with respect to individual attributes, multi-attribute fairness has not been systematically explored on this dataset, likely due to sample size constraints~\cite{Glaucoma2024,fami2025}. The cohort is racially balanced, enabling controlled analysis of group-specific performance.

In contrast, \textbf{MIMIC-CXR}~\cite{MIMIC-CXR2019} consists of frontal chest radiographs paired with free-text radiology reports. From this corpus, we derive a multi-label classification task over three clinically common pathologies—Pleural Effusion (PE - 25\% positive), Cardiomegaly (Car - 20\%), and Atelectasis (Act - 20\%)—spanning a broad range of demographic attributes. We include only patients with at least one of these conditions, yielding 82,282 images. This setting reflects a realistic multi-label diagnostic scenario in which all three pathologies are predicted jointly. Preprocessing and demographic subgroup definitions follow \cite{glocker2023algorithmic}. For both datasets, we treat age, race, and sex as sensitive attributes and use them to define demographic subgroups as described in Table \ref{tab:demographics_raw}, for evaluation.

\begin{table*}[t]
\centering
\caption{Sample size distribution across age, race, and sex subgroups for RNFL-OCT~\cite{Glaucoma2024} and MIMIC-CXR~\cite{MIMIC-CXR2019}, restricted to cases with PE, Car, or Act.}
\label{tab:demographics_raw}
\fontsize{8}{9.6}\selectfont 
\setlength{\tabcolsep}{4pt}
\renewcommand{\arraystretch}{1.15}
\begin{tabular}{l cccc ccc cc}
\toprule
\textbf{Dataset} & 
\multicolumn{4}{c}{\textbf{Age}} &
\multicolumn{3}{c}{\textbf{Race}} &
\multicolumn{2}{c}{\textbf{Sex}} \\
\midrule
\multirow{3}{*}{RNFL-OCT}
& \multicolumn{2}{c}{<60} & \multicolumn{2}{c}{>60}
& White & Black & Asian
& Female & Male \\
& \multicolumn{2}{c}{1,532} & \multicolumn{2}{c}{1,768}
& 1,100 & 1,100 & 1,100
& 1,812 & 1,488 \\
\midrule
\multirow{3}{*}{MIMIC-CXR} 
& 18--36 & 36--50 & 50--65 & 65+
& White & Black & Asian
& Female & Male \\
& 3,211 & 6,981  & 22,592  & 49,498 
& 65,858  & 13,196  & 3,228
& 37,310  & 44,972 \\
\bottomrule
\end{tabular}
\end{table*}

\subsection{Comparison with Existing Methods}
We compare against a standard cross-entropy based \textbf{Baseline} and five established fairness methods: \textbf{JTT}~\cite{JTT2021} (reweighting misclassified samples), \textbf{FairBatch}~\cite{fairbatch2021} (adaptive mini-batch balancing), \textbf{GDRO}~\cite{DRO2020} (worst-group reweighting), \textbf{FaMI}~\cite{fami2025} (mutual-information minimization), and \textbf{Adversarial Training (Adv)}~\cite{AdversarialLearning2018} (gradient-reversal-based debiasing).

\section{Experimental Setting}

All models are trained for 30 epochs on a single NVIDIA GeForce RTX 3090 GPU using a batch size of 64 and a learning rate of 0.001. Regularization coefficients are set to $\lambda_{\text{EO}^{+}}=\lambda_{\text{EO}^{-}}=0.5$ for RNFL-OCT and
$\lambda_{\text{EO}^{+}}=\lambda_{\text{EO}^{-}}=0.6$
for MIMIC-CXR, selected via validation-set tuning to balance overall fairness (Joint EO) and AUC. We use ResNet-34~\cite{he2016deep} for RNFL-OCT~\cite{Glaucoma2024} and DenseNet-121~\cite{huang2017densely} for MIMIC-CXR, and repeat all experiments over six independent random seeds. Following prior works, test time operating points are selected on the validation set maximising Youden’s J statistic~\cite{glocker2023algorithmic,gao2025fairread}.

\subsection{Evaluation Metrics}
We evaluate performance using AUC for overall accuracy and two fairness metrics~\cite{FairDisCo2023,XTranPrune2024}. EO disparity (\textbf{EOdds}) measures TPR/FPR disparity via the worst-to-best group ratio (lower is better), while \textbf{Equality of Opportunity across Multiple Subclasses (EOM)} captures per-class balance by averaging worst-to-best ratios across classes (higher is better)~\cite{xu2024addressing,du2022fairdisco}.

\begin{equation}
\text{EOdds} = 1 - \frac{1}{2}\left(\frac{\min_g \text{TPR}_g}{\max_g \text{TPR}_g} + \frac{\min_g \text{FPR}_g}{\max_g \text{FPR}_g}\right)
\label{eq:EO}
\end{equation}
\begin{equation}
\mathrm{EOM} = \frac{1}{K}\sum_{i \in \{0,1\}}
\frac{\min_{g}\Pr(\hat{y}=i \mid y=i,\, g)}
     {\max_{g}\Pr(\hat{y}=i \mid y=i,\, g)} ,
\end{equation}
where $K$ is the number of classes ($K{=}2$ for binary heads) and $g$ indexes
demographic subgroups of the attribute under evaluation. EOM is computed per attribute by taking the min and max over its subgroups.

\begin{table}[t]
\centering
\fontsize{8}{9.6}\selectfont
\setlength{\tabcolsep}{2.5pt}
\renewcommand{\arraystretch}{1.05}
\caption{Test-set performance across fairness methods. Values are reported as mean $\pm$ s.e.\; best and second best are shown in \textbf{bold} and \underline{underlined}. MIMIC-CXR results are macro-averaged across classes. \emph{Joint} denotes the average across Age, Race, and Sex, and $\Delta$AUC is computed relative to the dataset-specific baseline.}

\label{tab:results}
\resizebox{\linewidth}{!}{%
\begin{tabular}{l l cccc cccc c}
\hline
\multirow{2}{*}{\textbf{Dataset}} & \multirow{2}{*}{\textbf{Method}} &
\multicolumn{4}{c}{\textbf{EOdds ($\downarrow$)}} &
\multicolumn{4}{c}{\textbf{EOM ($\uparrow$)}} &
\multirow{2}{*}{\textbf{$\Delta$AUC ($\downarrow$)}} \\
\cmidrule(lr){3-6}\cmidrule(lr){7-10}
 & & Age & Race & Sex & Joint & Age & Race & Sex & Joint & \\
\hline
\multirow{7}{*}{RNFL-OCT}
 & Baseline   & 0.402\se{0.021} & 0.398\se{0.021} & 0.261\se{0.010} & 0.354\se{0.046}
 & 0.841\se{0.010} & 0.869\se{0.008} & 0.912\se{0.002} & 0.874\se{0.021}
 & --- \\
 & FairBatch  & 0.418\se{0.020} & 0.320\se{0.023} & 0.253\se{0.036} & 0.330\se{0.048}
 & 0.864\se{0.011} & 0.911\se{0.006} & 0.856\se{0.008} & 0.877\se{0.017}
 & -0.028\se{0.002} \\
 & GDRO       & 0.345\se{0.016} & 0.352\se{0.016} & 0.271\se{0.018} & 0.323\se{0.026}
 & \underline{0.877}\se{0.017} & \underline{0.930}\se{0.009} & 0.915\se{0.005} & \underline{0.907}\se{0.016}
 & -0.011\se{0.002} \\
 & JTT        & 0.314\se{0.028} & 0.335\se{0.030} & \underline{0.171}\se{0.012} & 0.273\se{0.052}
 & 0.855\se{0.019} & 0.894\se{0.007} & 0.899\se{0.006} & 0.883\se{0.007}
 & -0.061\se{0.008} \\
 & Adv        & 0.370\se{0.017} & \underline{0.304}\se{0.029} & 0.195\se{0.012} & 0.290\se{0.051}
 & 0.861\se{0.011} & 0.905\se{0.012} & \textbf{0.926}\se{0.003} & 0.897\se{0.019}
 & \underline{-0.008}\se{0.002} \\
 & FaMI       & \textbf{0.258}\se{0.031} & 0.325\se{0.034} & 0.173\se{0.026} & \underline{0.252}\se{0.044}
 & 0.834\se{0.020} & 0.871\se{0.012} & 0.916\se{0.010} & 0.874\se{0.024}
 & -0.061\se{0.009} \\
 & \textbf{Ours}
              & \underline{0.269}\se{0.016} & \textbf{0.249}\se{0.019} & \textbf{0.161}\se{0.018} & \textbf{0.226}\se{0.033}
              & \textbf{0.914}\se{0.012} & \textbf{0.939}\se{0.011} & \textbf{0.926}\se{0.002} & \textbf{0.926}\se{0.007}
              & \textbf{-0.006}\se{0.003} \\
\hline
\multirow{7}{*}{MIMIC-CXR}
 & Baseline   & 0.335\se{0.042} & 0.316\se{0.019} & 0.112\se{0.011} & 0.254\se{0.071}
 & 0.792\se{0.036} & 0.817\se{0.028} & 0.937\se{0.020} & 0.849\se{0.045}
 & --- \\
 & FairBatch  & 0.239\se{0.028} & 0.254\se{0.033} & 0.093\se{0.015} & 0.195\se{0.051}
 & 0.824\se{0.023} & 0.821\se{0.030} & 0.956\se{0.007} & 0.867\se{0.045}
 & -0.009\se{0.003} \\
 & GDRO       & 0.201\se{0.051} & 0.226\se{0.035} & 0.062\se{0.006} & 0.163\se{0.051}
 & \underline{0.884}\se{0.019} & 0.852\se{0.027} & 0.954\se{0.009} & 0.896\se{0.030}
 & \underline{-0.008}\se{0.004} \\
 & JTT        & \underline{0.192}\se{0.036} & 0.220\se{0.037} & 0.055\se{0.006} & 0.156\se{0.051}
 & 0.880\se{0.024} & 0.860\se{0.032} & \underline{0.963}\se{0.006} & \underline{0.901}\se{0.032}
 & -0.015\se{0.003} \\
 & Adv        & 0.216\se{0.036} & \underline{0.203}\se{0.054} & \textbf{0.044}\se{0.009} & \underline{0.154}\se{0.055}
 & 0.864\se{0.028} & \underline{0.871}\se{0.037} & 0.961\se{0.019} & 0.899\se{0.031}
 & -0.010\se{0.004} \\
 & FaMI       & 0.208\se{0.035} & 0.222\se{0.067} & 0.066\se{0.025} & 0.165\se{0.050}
 & 0.866\se{0.034} & 0.857\se{0.044} & 0.956\se{0.019} & 0.893\se{0.032}
 & -0.022\se{0.006} \\
 & \textbf{Ours}
              & \textbf{0.153}\se{0.019} & \textbf{0.174}\se{0.042} & \underline{0.054}\se{0.005} & \textbf{0.127}\se{0.037}
              & \textbf{0.897}\se{0.023} & \textbf{0.885}\se{0.035} & \textbf{0.966}\se{0.009} & \textbf{0.916}\se{0.025}
              & \textbf{-0.007}\se{0.005} \\
\hline
\end{tabular}%

}
\end{table}
\begin{figure}[h]
    \centering
    \includegraphics[width=\textwidth]{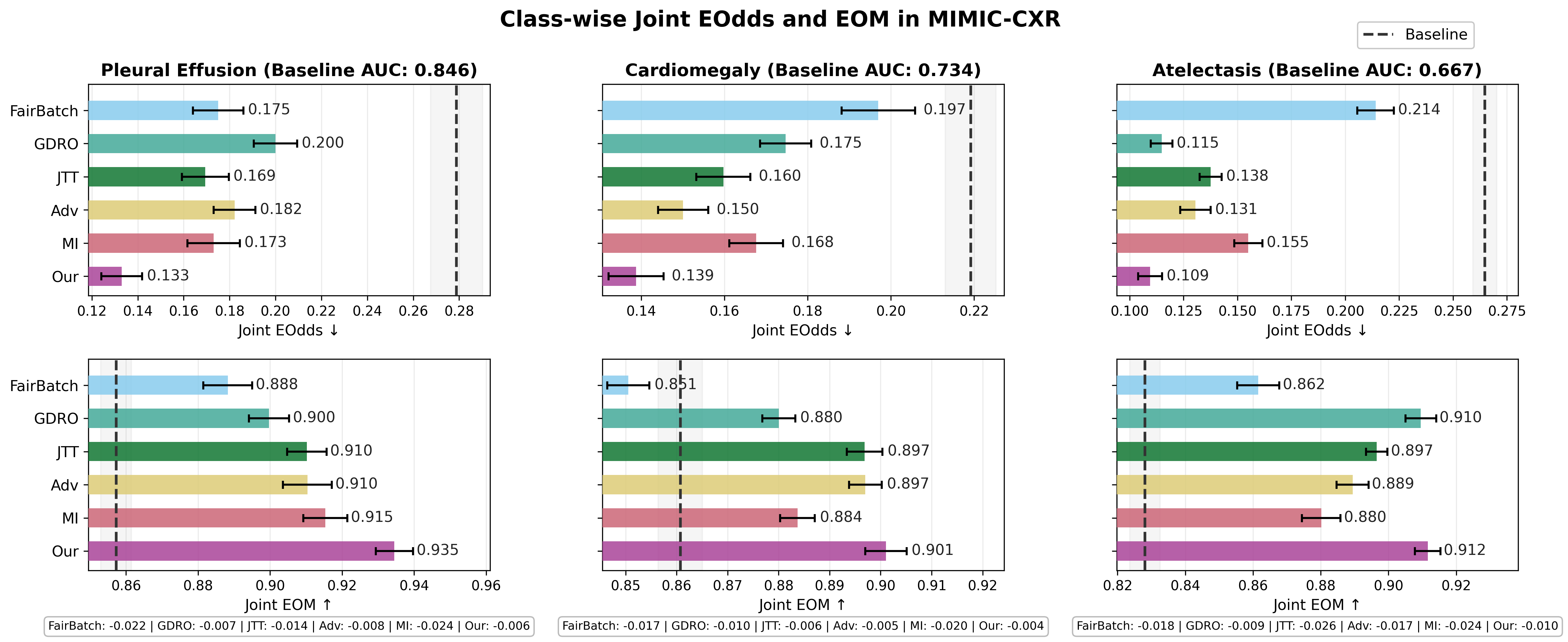}
    \caption{Class-wise Joint EOdds, EOM, and $\Delta$AUC (shown as box in bottom) on MIMIC-CXR. Error bars show standard error; grey shading denotes baseline standard error.}
    \label{fig2:main_results}
\end{figure}
\section{Results and Discussion}

Table~\ref{tab:results} summarizes performance across both datasets, with class-wise results for MIMIC-CXR shown in Fig.~\ref{fig2:main_results}. On RNFL-OCT, the baseline achieves the highest AUC (0.845) but exhibits substantial demographic disparities (joint EOdds = 0.354 and joint EOM = 0.874). Fairness-aware baselines reduce these gaps but typically incur noticeable accuracy degradation (e.g., FaMI $\Delta$AUC = -0.061). Our method substantially improves worst-group parity, achieving the lowest joint EOdds (0.226) and highest joint EOM (0.926) with a small AUC drop ($\Delta$AUC = -0.006). A similar pattern is observed on MIMIC-CXR. While the baseline attains the highest AUC (0.749), it suffers from significant disparities (joint EOdds = 0.254 and joint EOM = 0.849). Our approach reduces joint EOdds to 0.127 and increases joint EOM to 0.916, representing the strongest worst-group parity among all methods, with small discriminative performance drop ($\Delta$ AUC = -0.007). Compared to other mitigation strategies that either yield weaker parity improvements or incur larger accuracy losses, our method achieves a more favorable fairness–performance balance by mitigating localized AUC discrepancies in score regions that drive TPR/FPR disparities, resulting in minimal impact on overall AUC. 

We perform an ablation study on RNFL-OCT to assess the contributions of the two regularization components (Table~\ref{tab:ablation}). Activating either $\mathcal{L}_{\text{EO}^{+}}$ or $\mathcal{L}_{\text{EO}^{-}}$ alone yields partial fairness improvements, while enabling both achieves the strongest gains, confirming their complementary roles. Increasing $\lambda_{\text{EO}^{+}}$ and $\lambda_{\text{EO}^{-}}$ further improves fairness but leads to expected degradation in overall AUC~\cite{menon2018cost,mehrabi2021survey}.

\begin{table}[t]
\centering
\fontsize{8}{9.6}\selectfont
\setlength{\tabcolsep}{2.5pt}
\renewcommand{\arraystretch}{1.05}
\caption{Ablation study on the RNFL-OCT dataset, showing the individual impact of the regularization components $\mathcal{L}_{\text{EO}^{+}}$ and $\mathcal{L}_{\text{EO}^{-}}$.}
\label{tab:ablation}
\resizebox{\linewidth}{!}{%
\begin{tabular}{cc cccc cccc c}
\hline
\multirow{2}{*}{$\mathcal{L}_{\text{EO}^{+}}$} & \multirow{2}{*}{$\mathcal{L}_{\text{EO}^{-}}$} & \multicolumn{4}{c}{\textbf{EOdds ($\downarrow$)}} & \multicolumn{4}{c}{\textbf{EOM ($\uparrow$)}} & \multirow{2}{*}{\textbf{$\Delta$AUC ($\downarrow$)}} \\
\cmidrule(lr){3-6}\cmidrule(lr){7-10}
& & Age & Race & Sex & Joint & Age & Race & Sex & Joint & \\
\hline
$\checkmark$ & $\times$ & 0.297\se{0.020} & 0.275\se{0.025} & 0.192\se{0.015} & 0.255\se{0.020} & 0.901\se{0.015} & 0.915\se{0.010} & 0.913\se{0.008} & 0.910\se{0.011} & -0.006\se{0.001} \\
$\times$ & $\checkmark$ & 0.262\se{0.018} & 0.266\se{0.022} & 0.217\se{0.012} & 0.248\se{0.018} & 0.885\se{0.013} & 0.891\se{0.012} & 0.902\se{0.006} & 0.893\se{0.011} & -0.005\se{0.001} \\
\hline
\end{tabular}%
}
\end{table}

\section{Conclusion}
We propose a worst-group equalized-odds margin regularizer for multi-label medical image classification that addresses a critical need to mitigate worst-case subgroup disparities across multiple demographic attributes using differentiable log-sum-exp margins. Our formulation directly penalizes localized score discrepancies responsible for over- and under-diagnostic fairness violations with serious clinical consequences, while preserving diagnostic performance. Finally, we provide a unified mechanism that replaces traditional collections of attribute-specific constraints with a single training objective.
\section*{Acknowledgements}
This work was supported by GSK AI through the Responsible AI Research Grant. The authors also acknowledge the Australian Institute for Machine Learning for providing institutional support and research infrastructure.
\bibliography{references}
\bibliographystyle{splncs04}

\end{document}